\documentclass[11pt]{article}
\usepackage{colacl}
\usepackage[dvips]{graphicx}
\usepackage{e-narrowitem}
\usepackage{e-narrowcaption}
\title{Analysis of Titles and Readers\\
-- For Title Generation Centered on the Readers}
\author{Yasuko Senda\dag\ddag \and Yasusi Sinohara\dag\\
\dag Communication \& Information Research Laboratory\\
Central Research Institute of Electric Power Industry, Japan\\
\ddag Department of Computational Intelligence and Systems Science\\
Tokyo Institute of Technology, Japan\\
\{senda, sinohara\}@criepi.denken.or.jp
}
\begin{document}
\maketitle
\begin{abstract}
The title of a document has two roles, to give a compact summary and to
 lead the reader to read the document.
Conventional title generation focuses on finding key expressions from
 the author's wording in the document
to give a compact summary and pays little attention to the reader's
 interest.
To make the title play its second
role properly, it is indispensable to clarify the content (``what to say'') and wording (``how to say'') of titles
that are effective to attract the target reader's interest.
In this article, we first identify typical content  and wording of titles
aimed at general readers in a comparative study between titles of technical papers
and headlines rewritten for newspapers. Next, we describe the results of
a questionnaire survey on the effects of the content and wording of titles
on the  reader's interest.
The survey of general and knowledgeable readers shows both common and different tendencies in interest.
\end{abstract}

\section{Introduction}

The title is expected to play two roles.
One is to give the reader a very compact summary of the document,
 and 
the other is to attract the target reader's interest
and lead the reader to read the document.
It is preferable that a title plays both roles,
because the reader may be disappointed with a gap between the title
and the document if the title plays the former role poorly, and
the reader may not read the document 
 if the title plays the latter role poorly.
Therefore, it is very important what title is attached to a document.

Several techniques have been reported on generating titles \cite{Jin}\cite{Berger},
and they focus on the former role, that is, to give a compact summary.
The main approach is to find a few 
keywords from the document
by calculating the importance of each word in the document.
This approach, incidentally, is similar to most text summarization techniques.
The selected keywords or title strongly reflect the author's wordings.
In other words, this approach is an ``author-centered approach''.
In some cases, the title generated by this approach might play the latter role poorly
and fail to get the reader's interest.

To make generated titles play their latter role properly,
it is not sufficient to look into only the author's document.
It is important to also pay more attention to the reader.
It is necessary in the ``reader-centered approach''
to clarify the features of the reader's attention, that is, 
the relationship between  the reader's attention and the content and
wording of the title.
Based on this knowledge, it will be possible to extract information from the document
that is more attractive to the reader than the author's key expressions
and to include it in the generated title. 

Our first goal is to clarify the kind of content and wording that are key
to getting the reader's comprehension,
interest in, and positive feeling toward the document.
For this purpose, we first conducted a comparative study on the titles of technical papers
 and headlines rewritten for newspapers
 to identify the typical content and wording of titles aimed at general readers.
We then conducted a questionnaire survey on the effects of different
content and wording of titles on different kinds of readers' interest.

The following sections are as follows:
section \ref{comparison} describes the comparative study and its results,
section \ref{quest} and \ref{differinImpre} explain the details and the
results of our questionnaire survey,
and section \ref{discuss} relates our conclusion and future work.

\section{Comparison between Paper Titles and Newspaper Headlines}
\label{comparison}
To identify the typical content and wording of titles aimed at general readers,
we conducted a comparative study on the Japanese titles of research papers and
the headlines of Japanese newspaper articles describing the corresponding technology.

We first divided the titles and headlines into several syntactic components (text segments)
and assigned each component a syntactic function tag,
such as ``P(urpose of development)'' and  ``M(ethod for realization)'',
and then comparatively analyzed the expression of paper title and headlines
on the basis of each component.

\subsection{Overview of Collected Documents}
\begin{table*}[tbhp]
\small
\begin{center}
\caption{Features of Documents Used in the Comparison of Their Titles or Headlines}
\label{AttrTitle}
\begin{tabular}{|c|c|c|c|c|c|c|}    \hline
  & \multicolumn{2}{|c|}{\bf Kind of} & {\bf Target} & {\bf Author}
  & {\bf  Publication} & {\bf Number of} \\
  & \multicolumn{2}{|c|}{\bf Documents} & {\bf Reader} &   & {\bf Date}
 & {\bf Documents} \\ \hline
1 & \multicolumn{2}{|c|}{Article} & Researcher & Researcher & '89--'00 & 1464 \\ \hline
2 & \multicolumn{2}{|c|}{Technical Report} & General & Researcher & '80--'00 & 2861 \\ 
  & \multicolumn{2}{|c|}{¡¡} &  Reader &  &  &  \\ \hline
3 & Trade & on the Power Industry &  &  &  & 531 \\ \cline{1-1} \cline{3-3} \cline{7-7}
4 & Newspaper & on Technology & General & Newspaperman & '88--'98 & 313 \\ \cline{1-1} \cline{3-3} \cline{7-7}
5 &  & on the Economy & Reader &  &  & 203 \\ \cline{1-1} \cline{2-3} \cline{7-7}
6 & \multicolumn{2}{|c|}{General Newspaper} &  &  &  & 364 \\ \hline
\end{tabular}
\end{center}
\end{table*}%
Table \ref{AttrTitle} shows the outlines of collected documents, including research papers
(articles and technical reports) and newspaper articles
 (from three trade newspapers and one general newspaper).
The Japanese titles and headlines were retrieved from a document database at a research institute.
They cover science and technology at large.

We identified one or a few technical papers closely related to about
90\% of the headlines.

\subsection{Syntactic Function Tags}

\begin{table*}[thbp]
\small
\begin{center}
\caption{Components Included in the Title of Technical Papers}
\label{TitlePattern}
\begin{tabular}{|l|p{4.5cm}|p{2.6cm}|p{5cm}|p{1.7cm}|}
\hline
  & {\bf Each Syntactic Component's Function} & {\bf Syntactic Category}
  & {\bf Specific Expression or Preposition} & {\bf Frequency} \\ \hline
{\bf B} & Behavior of the Technology & Verbal Noun & - & $> 80\%$ \\ \hline
{\bf O} & Object of the Behavior & Noun Phrase & Japanese particle
 ``{\it wo}''  &  $> 80\%$\\ \hline
{\bf M} & Method for Realizing & Noun Phrase & Japanese expression
 ``{\it ni yori}'', etc. & $< 30\%$ \\
 & & & (such as ``by'' in English) & \\ \hline
{\bf S} & Strong Point & {Adjective/Adverb Phrase} & - & $< 30\%$\\ \hline
{\bf P} & Purpose of Development & Noun Phrase & Japanese expression ``{\it no tame ni/no}'', etc. & $< 30\%$\\ 
 & & & (such as ``for'' in English) & \\ \hline
{\bf T} & Technology type & Noun & - & $58\%$\\ \hline
{\bf D} & Development & Noun & Japanese expression ``{\it no kaihatsu}'', etc. & \\
 & & & (such as ``development of'')  & \\ \hline
\end{tabular}
\end{center}
\end{table*}%

We found that most text segments of the collected titles and headlines
could be classified into the following syntactic function tags.
Here, we define each syntactic function tag and describe the tagging process we used in the analysis.

\begin{itemize}
\item {\bf B}(ehavior):
Behavior is expressed by a verbal noun or verb.
For example, in the title ``Development of an Exploration System of Buried Cables'',
the word ``exploration'' describes the main behavior or action of the developed technology
and is assigned the tag ``B''.
\item {\bf O}(bject):
The object is a noun phrase in the objective case of the verbal noun of behavior ``B''.
In the above example, the compound noun ``buried cables'' is assigned
      the tag ``O''.
\item {\bf T}(echnology type):
The words assigned the tag ``T''(echnology type) are very restricted
and include such words as ``system'', ``model'' and ``method''.
They express the form of the developed technology. 
\item {\bf P}(urpose):
The purpose of development is usually expressed by a noun phrase following a preposition,
such as ``for'' (``{\it no tame-ni/no}'' in Japanese).
(e.g., ``for power distribution cables under pavements'')
\item {\bf M}(ethod):
The tag ``M'' stands for the method for realization
and  is expressed by a noun phrase following a preposition,
such as ``by'' (``{\it ni yori}'' in Japanese).
(e.g., ``by underground radar'')
\item {\bf S}(trong point):
``S'' stands for the advantage or merit of the developed technology.
It is usually expressed by an adjective or adverb phrase.
(e.g., ``highly accurate'').
\item {\bf  D}(evelopment):
``D'' stands for the expression that frequently appears in paper titles and headlines,
such as ``the development of'', ``the evaluation of'', and ``a study on''.
\item {\bf E}(t alia):
There are several text segments that do not fall into the above seven categories.
They appear mainly in newspaper headlines.
Two examples are names of development organizations and descriptions of the issue for practical use of the technology.
\end{itemize}

We excluded the two tags ``D'' and ``E'' in a comparative analysis
because they do not directly relate to the content of the developed technology.

\subsection{Tagging Process}

As for paper titles (in Japanese),
we divided them semiautomatically according to the following pattern
\begin{center}
{\small P? M? S? O S? B S? T? D? \\}
{\small (Each alphabet means syntactic function tag.\\
``?'' means ``either zero times or one time''.)
}
\end{center}
using the ``ChaSen'' Japanese morphological analyzer \cite{chasen} and
syntactic clues (such as word order, syntactic category and specific prepositions).
The results are reviewed and errors are manually corrected.

As for headlines, automatic tagging is difficult because of
the frequent use of verbal omissions and inversion.
Therefore, we manually divided the headlines.
In order to improve the precision of human tagging as highly as possible,
the human tagger divided them according to the following procedure.
\begin{enumerate}
\item Find the verbs (or verbal nouns).
\item Determine the verb (or verbal noun) which corresponds to ``B''
using syntactic and semantic clues (such as special prepositions and
 semantic modification relations).
\item Identify the other components which modify ``B''
using the clues used in the above step 2.
\end{enumerate}

\subsection{Comparative Analysis}
We analyzed the results of the above tagging in the following two ways.
\begin{itemize}
\item [1.] The difference in the frequency of the six main components with tag 
B(ehavior), O(bject), M(ethod), S(trong point),
P(urpose) and T(echnology type) in paper titles and headlines.
\item [2.] The difference in the expression of the same syntactic components in paper titles and headlines.
\end{itemize}

\subsection{Obligatory and Optional Components}
The last column in Table \ref{TitlePattern} indicates the frequency of
the occurrence of each tag in titles and headlines.
The components (text segments) with tags ``B'' , ``O'', and ``T'' appear
more than 80\%, 80\%, and 58\%, respectively, in titles and headlines
\footnote{``B'' and ``O'' appear 100\% in the paper titles and 80\% in the
headlines because verbal omissions sometimes occur in newspaper headlines.}
.
We call these high-frequency components ``obligatory component'' and 
the other components with ``P'', ``M'', and ``S'', ``optional components''
.

\subsubsection{Obligatory Components}
Obligatory components are important
because they are essential in reporting the substance of the technology to the reader.



By comparing the expression of the obligatory components (labeled ``T'', ``B'', and ``O''),
we categorized the difference of the expression into two points.
One is that technical jargons are used in paper titles while plain terms are
used in headlines to express the same thing.
The other is that 
 in the obligatory component
 the title explains the concrete content of the technology
while the headline expresses the purpose of the technology,
which general readers can understand.


In other words, we identified two expressing techniques of newspapermen.
\begin{itemize}
\item Instead of using technical jargons, the plain synonymous terms are used.
\item Instead of expressing 
what the technology does, 
they express what the purpose of the technology is.
\end{itemize}

An example of the former in English is as follows:
{\small 
\begin{description}
\item [Title:] Method to Shorten Radioactive Half-life
\item [Headline:] Method to Shorten the Duration of Radiation
\end{description}
}

An example of the latter is as follows:
{\small 
\begin{description}
\item [Title:] Method to Shorten Radioactive Half-life
\item [Headline:] Method to Shorten Storage Period of Radioactive Waste
\end{description}
}

The expression patterns typical in obligatory components are summarized in Table \ref{tableHissu}.
It is arranged from the two viewpoints of ``what to say'' and ``how to say''.
\begin{table*}[bthp]
\small
\begin{center}
\caption{Expression Patterns in Obligatory Components}
\label{tableHissu}
\begin{tabular}{|l||p{38mm}|p{45mm}|p{38mm}|}    \hline
 \multicolumn{1}{|c||}{What to} &  \multicolumn{2}{c|}{\bf Pattern 1} &
 \multicolumn{1}{c|}{\bf Pattern 2} \\
 \multicolumn{1}{|c||}{say} &  \multicolumn{2}{c|}{(What the technology
 does)} & \multicolumn{1}{p{38mm}|}{(What the purpose of the technology is)} \\ \hline
 \multicolumn{1}{|c||}{How to} &  \multicolumn{1}{c|}{\bf Pattern 1.1} &
 \multicolumn{1}{c|}{\bf Pattern 1.2} & \multicolumn{1}{c|}{\bf Pattern 2.0} \\
 \multicolumn{1}{|c||}{say} &  \multicolumn{1}{c|}{(Using technical jargon)} &
 \multicolumn{1}{c|}{(in plain terms)} & \multicolumn{1}{c|}{(in plain terms)} \\ \hline \hline
T(ype) & Method & Method & Method \\ \cline{1-1}
B(ehavior) & to  Shorten &  to Shorten  &  to Shorten  \\ \cline{1-1}
O(bject) & {\bf Radioactive Half-life} & {\bf the Duration of Radiation}
 & {\bf Storage Period of Radioactive Waste} \\ \hline
 \end{tabular}
\end{center}
\end{table*}

\subsubsection{Optional Components}


By comparing the expression of the optional components,
in the ``M'' component
we found that technical jargons are frequently used in paper titles while plain terms are
used in headlines.
Moreover, 
by analyzing the frequency of component combinations,
we found that the combination of optional components with ``M'' and ``S'' is
seldom used in any title or headline.
We also found that the use of ``M'' in titles is more frequent
than in headlines
while  the description of the advantages (``S'') in headlines is
five times more frequent than in titles.
The analysis suggests that the description of ``M'' found in paper titles are omitted
in headlines and the description of advantages (``S'') are used in headlines.

The expressing techniques we identified are as follows:
\begin{itemize}
\item Instead of using technical jargons, the plain synonymous terms are
       used in the ``M'' component.
\item Instead of expressing the method of realizing the technology,
they express the advantage (``S'') of the technology.
\end{itemize}

An example of the former in English is as follows.
{\small 
\begin{description}
\item [Title:] Method to Shorten Storage Period of Radioactive Waste by Metallic Fuel FBR
\item [Headline:] Method to Shorten Storage Period of Radioactive Waste by Burnout
\end{description}
}

An example of the latter is as follows:
{\small 
\begin{description}
\item [Title:] Method to Shorten Storage Period of Radioactive Waste by Metallic Fuel FBR
\item [Headline:] Method to Shorten Storage Period of Radioactive Waste by 1/10000
\end{description}
}

The expression patterns typical in optional components are summarized in Table \ref{tableNini}.
The table is also arranged from the two viewpoints of ``what to say'' and ``how
to say'', as in Table \ref{tableHissu}.
\begin{table*}[bthp]
\small
\begin{center}
\caption{Expression Patterns in Optional Components}
\label{tableNini}
\begin{tabular}{|l||p{38mm}|p{35mm}|p{38mm}|}    \hline
 \multicolumn{1}{|c||}{What to} &  \multicolumn{2}{c|}{\bf Pattern 3} &
 \multicolumn{1}{c|}{\bf Pattern 4} \\
 \multicolumn{1}{|c||}{say} &  \multicolumn{2}{c|}{(What the method of
 realizing the technology is)} & \multicolumn{1}{p{38mm}|}{(What the strong point of the technology is)} \\ \hline
 \multicolumn{1}{|c||}{How to} &  \multicolumn{1}{c|}{\bf Pattern 3.1} &
 \multicolumn{1}{c|}{\bf Pattern 3.2} & \multicolumn{1}{c|}{\bf Pattern 4.0} \\
 \multicolumn{1}{|c||}{say} &  \multicolumn{1}{c|}{(Using technical jargon)} &
 \multicolumn{1}{c|}{(in plain terms)} & \multicolumn{1}{c|}{(in plain terms)} \\ \hline \hline
T(ype) & Method & Method & Method \\ \cline{1-1}
B(ehavior) & to  Shorten &  to Shorten  &  to Shorten  \\ \cline{1-1}
O(bject) & Storage Period of Radioactive Waste & Storage  Period of Radioactive Waste & Storage Period of Radioactive Waste \\ \cline{1-1}
M(ethod) & {\bf by Metallic Fuel FBR} &  {\bf by Burnout}  &  \\ \cline{1-1}
S(trong Point) &  &    & {\bf by 1/10000}  \\ \hline
\end{tabular}
\end{center}
\end{table*}

\section{Questionnaire Survey}
\label{quest}
In this section, we explain the details of our questionnaire survey.

\subsection{Viewpoint of the Survey}

We assume that
the factors which affect the reader's impression of a title which
expresses newly developed technology are the following:
\begin{description}
\item[F1] Mode of expression for titles (in other word, the expression patterns)
\item[F2] Whether or not the reader has an interest in the technical field
\item[F3] The reader's level of expertise in the technical field
\end{description}

Therefore, in order to clarify
what kind of impression the title based on each expression pattern (F1) gives the reader,
we should investigate how the impressions
that each reader who differs in F2 and F3 receives
change with each expression pattern.
We then prepared our questionnaire survey according to the following procedures.
\begin{enumerate}
\item Select the technical fields for the survey (we selected three
      fields: electric transmission, architectural engineering, and environmental science).
\item Categorize the respondents by their interest in each
      technical field and their expertise in that field.
\item Generate titles describing new technologies in the three technical fields
by using expression patterns
\item Draw up a questionnaire which asks the respondents 
what kind of impression they had when they read each generated title.
\end{enumerate}

In the following three subsections, we explain the details of the above
procedures step 2, 3 and 4, in order.

\subsection{Categorization of Respondents}
In order to categorize the respondents by their interest in each
      technical field and their expertise of that field,
we asked the respondents two questions
in the preliminary questionnaire.

One asked them if they were concerned/unconcerned with one of the
three technical fields, and the other asked them about their main source of information
(three options:
general newspaper, trade newspaper, and academic journal,
were given,
if they were concerned).

We categorized them into two types (``Unconcerned'' and ``Concerned'')
by the former question.
We then categorized ``Concerned'' into three types (``Commoner'',
``Engineer'' and  ``Researcher'', whose main sources of information are
general newspaper, trade newspaper and academic journal in turn.) by the latter question.
We labeled ``Unconcerned'' and three types of ``Concerned''
(``Commoner'', ``Engineer'' and  ``Researcher'') as readership.
Table \ref{kaishu3} shows the number of respondents by readership.
\begin{table*}[htbp]
\small
\begin{center}
\caption{Number of Respondents}
\label{kaishu3}
\begin{tabular}{|l|r|r|r|r|}    \hline
\it  & \multicolumn{1}{|l|}{Unconcerned} & \multicolumn{3}{|c|}{Concerned} \\ \cline{3-5}
\it  &  &
 \multicolumn{1}{|l|}{Commoner} &
 \multicolumn{1}{|l|}{Engineer} & \multicolumn{1}{|l|}{Researcher} \\ \hline
\it {\bf Electric Transmission} & 151 & 114 & 69 & 48 \\ \hline
\it {\bf Architectural Engineering} & 168 & 115 & 62 & 45 \\ \hline
\it {\bf Environmental Science} & 108 & 153 & 71 & 53 \\ \hline
\end{tabular}
\end{center}
\end{table*}%

\subsection{Titles Prepared for Our Questionnaire}

In this section,
we explain the method for composing the titles used in the questionnaire.
The procedure for composing the titles follows.
\begin{enumerate}
\item Select three new technologies in each of the three technical fields, which were
developed in a research institute.
\item Retrieve the phrases of titles and headlines concerning each technology,
 from the document database of the institute, that
correspond to each expression pattern.
\item Combine the phrases to compose twelve titles per each technology,
using three types of obligatory component expression pattern of
by four types of optional component expression pattern
 (including 
a pattern without optional components). The total number of the titles per one
field is 36 (twelve titles by three fields).
\end{enumerate}

We divided thirty six titles per each field into four groups (each group
consists of nine titles).
The breakdown of the nine titles is three types of obligatory component
expression pattern 
 by three types of optional component expression pattern.
We randomly divided the respondents of each readership into four groups
and then asked four groups of the respondents about the impression of
four groups of the titles respectively.

\subsection{Contents of Our Questionnaire}
In order to investigate whether each prepared title is effective in getting
each readership's comprehension, positive feelings, and interest,
we asked the respondents the following questions in turn.
\begin{itemize}
\item Do you think the title is comprehensible?
\item Do you feel positive toward the technology after reading this?
\item Do you want to know more about the technology?
\end{itemize}

\subsection{Method}
Respondents to the questionnaire
were the staff of a research institute and 
monitors of a marketing research firm.
We asked them to answer our questionnaire on a Web page
which is prepared for this survey.

\section{Difference in Impression According to 
Obligatory Component Expression Pattern}
\label{differinImpre}
\begin{table*}[htbp]
\small
\begin{center}
\caption{Results of the Chi-square Test and Cramer's V}
\label{test}
(In this table, ``Chi-square'' means the significant 
level tested by the
 Chi-square, \\
and ``Cramer's'' means the value calculated by Cramer's
 V.)
\begin{tabular}{|r|r|r|r|r|r|}    \hline
\it  & & \multicolumn{1}{|l|}{\bf Unconcerned} & \multicolumn{3}{|c|}{\bf Concerned} \\ \cline{4-6}
\it  & & & \multicolumn{1}{|l|}{\bf Commoner} & \multicolumn{1}{|l|}{\bf Engineer} & \multicolumn{1}{|l|}{\bf Researcher} \\ \hline
\multicolumn{1}{|l|}{\bf Comprehensible} & \multicolumn{1}{l|}{\bf Chi-square} & 1\% & 1\% & 1\% & 1\% \\ \cline{2-6}
  & \multicolumn{1}{|l|}{\bf Cramer's} & $0.62$ & $0.59$ & $0.38$ &$0.43$ \\ \hline
\multicolumn{1}{|l|}{\bf Able to Evoke} & \multicolumn{1}{l|}{\bf Chi-square} & 1\% & 1\% & 1\% & 1\% \\ \cline{2-6}
\multicolumn{1}{|l|}{\bf Positive Feelings} & \multicolumn{1}{|l|}{\bf Cramer's} & $0.45$ & $0.45$ & $0.22$ & $0.24$ \\ \hline
\multicolumn{1}{|l|}{\bf Interesting} & \multicolumn{1}{l|}{\bf Chi-square} & 1\% & 1\% & 1\% & 5\% \\ \cline{2-6}
  & \multicolumn{1}{|l|}{\bf Cramer's} & $0.27$ & $0.35$ & $0.22$ & $0.1$ \\ \hline
\end{tabular}
\end{center}
\end{table*}%
\begin{figure}[bhtp]
\includegraphics[width=80mm]{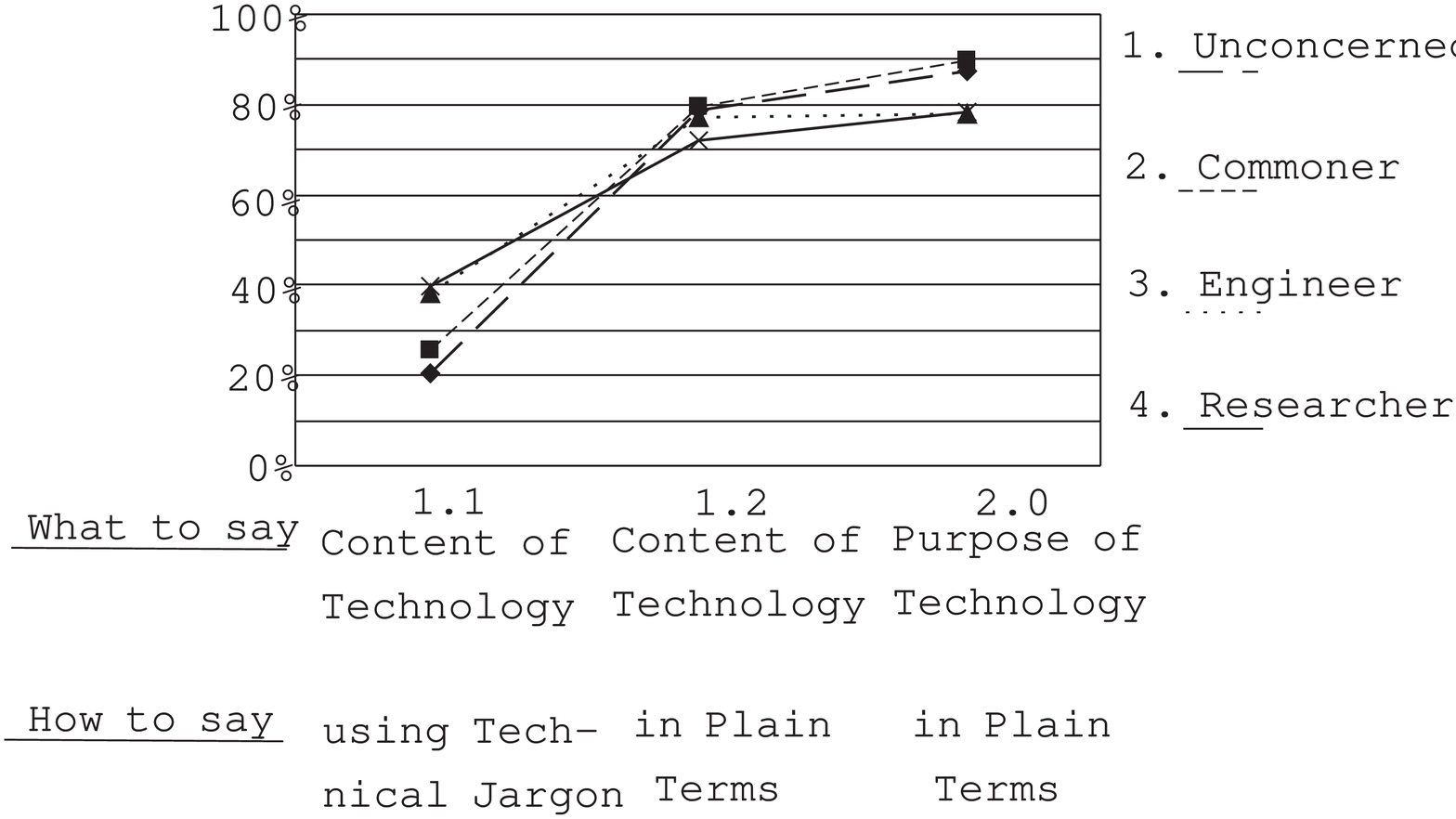}
\caption{Percentage of Titles That Respondents Regard as Comprehensible}
\label{rikai}
\end{figure}%
\begin{figure}[bhtp]
\includegraphics[width=80mm]{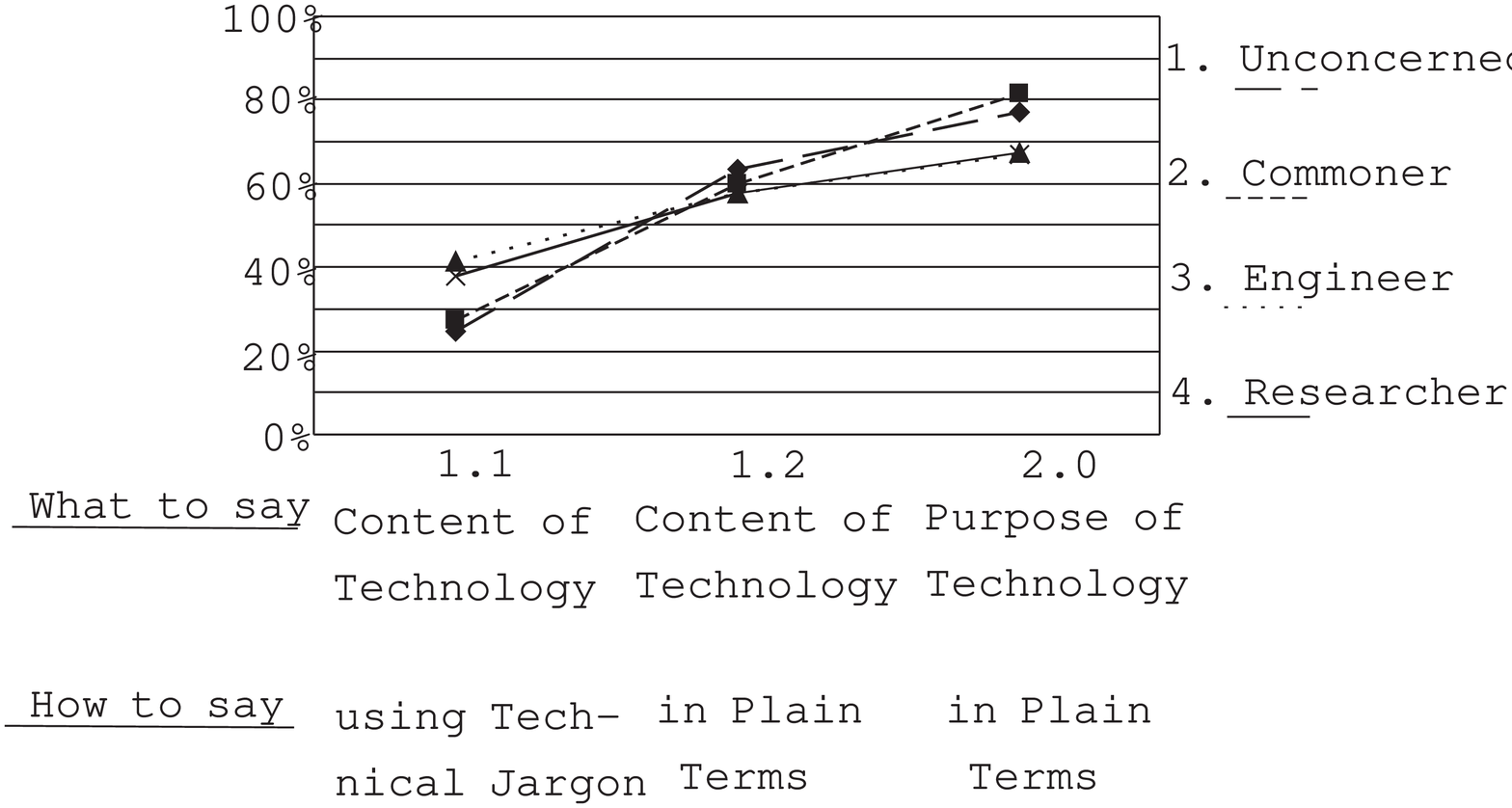}
\caption{Percentage of Titles That Respondents Regard as Able to Evoke Positive Feelings}
\label{good}
\end{figure}%
\begin{figure}[bhtp]
\includegraphics[width=80mm]{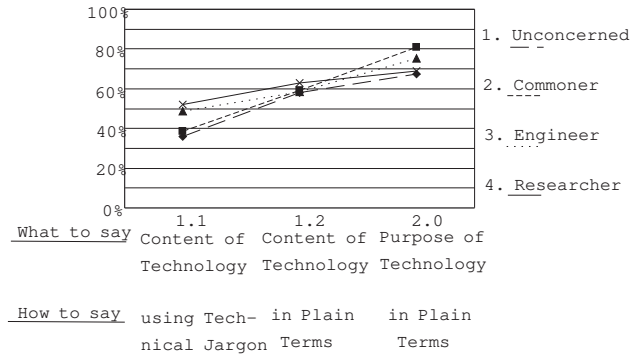}
\caption{Percentage of Titles that Respondents Regard as Interesting}
\label{kansin}
\end{figure}%

In this section, we report on the analysis of the 
correlation between
the obligatory component expression pattern in titles
and each readership's impression of the titles.

Figure \ref{rikai} shows the graph of the percentage of
 titles regarded as ``comprehensible'' by the respondents with 
each readership using each expression pattern. 
Figure \ref{good} and \ref{kansin} also show the ones 
regarded as ``evoked positive feelings'', ``interesting'' 
respectively.
We also made a 3x2 contingency table (expression 
patterns $\times$ yes/no answer on impressions) for 
each impression and each readership
and calculated the significance level of the $\chi^2$-test
 and the Cramer's V
\footnote{$V = \sqrt{X2 / N(k-1)}$
 where N $=$ total number of cases in the table; k $=$ 
number of
rows or columns, whichever is smaller.
 Cramer's V ranges 0 to 1 where 0 means that the two 
value variables are perfectly unrelated
and 1 means that they are perfectly related.}.

Because all significance levels in Table \ref{test} are
 under 5\%,
it is confirmed that there exists associations among 
expression patterns and impressions
for each readership.

However, the strength of the associations is 
low
as for the respondents with readership ``engineer'' or
``researcher'',
who have expertise, because of lower values of Cramer's
 V,
while the strength of the association is 
high
 as for those who has readership ``unconcerned'' or
``commoner''
because of higher values of V.

Because all graphs slope up from left to right
in Figures \ref{rikai}, \ref{good} and \ref{kansin},
the expression pattern 2.0 (expressing what the purpose 
of the technology is)
is the most effective to make a title impressive
(more precisely, regarded as ``comprehensible'', ``evoked 
positive feelings''
and ``interesting'') for readers with most of readership,
 especially
the ``unconcerned'' readers or ``commoner''.
As for the readers with expertise, other types of 
expression patterns such as 1.1 or 1.2
should also be considered because of their flatter 
slopes.

\section{Conclusion and Future Work}
\label{discuss}

We emphasized the need for title generation centered on the reader
and identified the typical content and wording of titles aimed at general
readers
by conducting a comparative study on paper titles and headlines.
Moreover,
we verified the effects of different content and wording
of titles.
As a result,
titles using expression pattern 2.0 (expressing 
what the purpose of the technology is) in obligatory component is the most effective
in getting the general reader's comprehension, positive feelings, and interest.
However,
the more expertise in a technical field the reader has,
the less the reader tends to be influenced.

In future work, we will verify the difference in impression according to optional
component expression patterns
 by analyzing the results of our questionnaire survey.
We plan to establish the method for generating a title by extracting
phrases from the body text and combining them, 
which correspond to the expression pattern which is effective in getting
target reader's interest.

\vspace{1mm}

\begin{flushleft}
{\bf Acknowledgements}
\end{flushleft}
The authors would like to express our gratitude to Dr. Manabu Okumura of
Tokyo Institute of Technology for his valuable suggestions to improve
our research and
anonymous reviewers for their suggestions to improve our paper.

\small
\bibliographystyle{acl}  
\bibliography{coling}
\end{document}